\documentclass[10pt,twocolumn,letterpaper]{article}

\usepackage{iccv}
\usepackage{times}
\usepackage{epsfig}
\usepackage{graphicx}
\usepackage{amsmath}
\usepackage{amssymb}
\usepackage{multirow}
\usepackage{enumitem}
\usepackage{marvosym}
\usepackage{fancyhdr}
\setitemize[1]{itemsep=0pt,partopsep=0pt,parsep=\parskip,topsep=0pt}

\usepackage[pagebackref=true,breaklinks=true,letterpaper=true,colorlinks,bookmarks=false]{hyperref}

\iccvfinalcopy 

\def\iccvPaperID{6791} 

\ificcvfinal\fi

\begin{document}


\title{HandFoldingNet: A 3D Hand Pose Estimation Network Using Multiscale-Feature Guided Folding of a 2D Hand Skeleton}











\author{Wencan Cheng\textsuperscript{1},  Jae Hyun Park\textsuperscript{1},  Jong Hwan Ko\textsuperscript{2}\footnotemark[1]\\
\textsuperscript{1}Department of Artificial Intelligence, Sungkyunkwan University\\
\textsuperscript{2}College of Information and Communication Engineering, Sungkyunkwan University\\
{\tt\small \{cwc1260, xoxc4565, jhko\}@skku.edu}
}

\maketitle
\ificcvfinal\thispagestyle{fancy}\fi
\lhead{Accepted as a conference paper at International Conference on Computer Vision (ICCV) 2021}
\rhead{}
\cfoot{}
\renewcommand{\thefootnote}{\fnsymbol{footnote}} 
\footnotetext[1]{Jong Hwan Ko is the corresponding author.}

\begin{abstract}
With increasing applications of 3D hand pose estimation in various human-computer interaction applications, convolution neural networks (CNNs) based estimation models have been actively explored. However, the existing models require complex architectures or redundant computational resources to trade with the acceptable accuracy. To tackle this limitation, this paper proposes HandFoldingNet, an accurate and efficient hand pose estimator that regresses the hand joint locations from the normalized 3D hand point cloud input. The proposed model utilizes a folding-based decoder that folds a given 2D hand skeleton into the corresponding joint coordinates. For higher estimation accuracy, folding is guided by multi-scale features, which include both global and joint-wise local features. Experimental results show that the proposed model outperforms the existing methods on three hand pose benchmark datasets with the lowest model parameter requirement. 
Code is available at \url{https://github.com/cwc1260/HandFold}.


\end{abstract}

\section{Introduction}




3D hand pose estimation aims to estimate joint locations from input hand images. 
Accurate and real-time estimation is critical in various human-computer interaction applications, especially in virtual reality and augmented reality \cite{li2019survey, erol2007vision, marchand2015pose}.
Recently, many studies achieved impressive progress by utilizing hand depth images from depth cameras. 
However, it still remains challenging to achieve accurate and real-time estimation, due to various issues such as self-occlusion, noise, high dimensionality, and various orientations of a hand \cite{ge2018point, ge2018hand, moon2018v2v, du2019crossinfonet}.

\begin{figure}
\centering
\includegraphics[width=8cm]{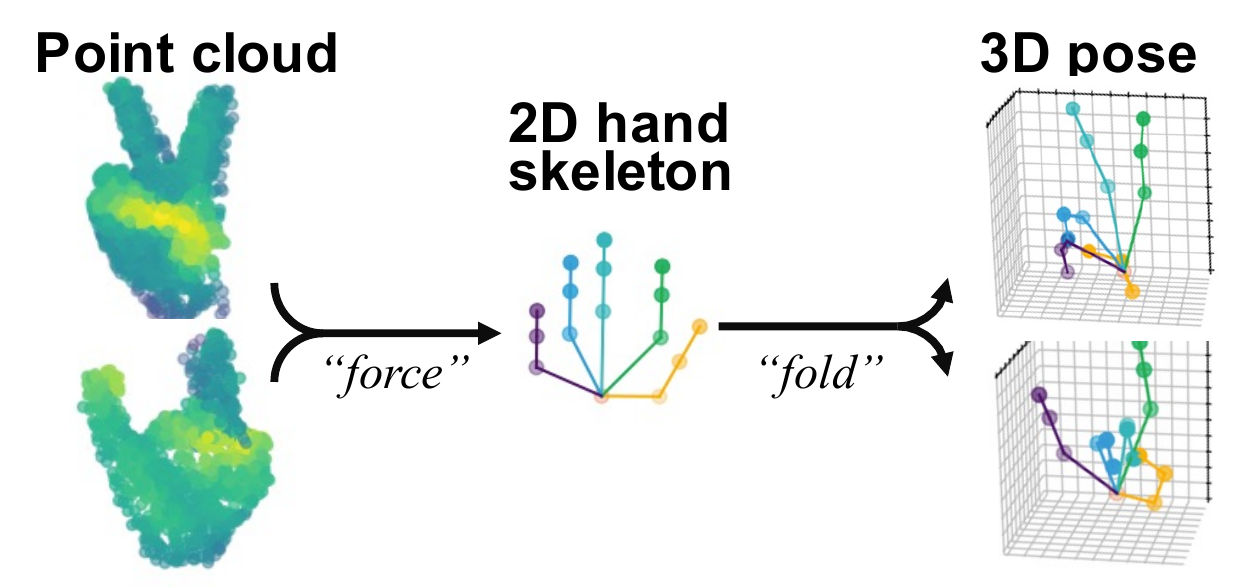}
\caption{Illustration of the folding concept. The network can be interpreted as emulating the "force" through multi-scale features extracted from the point cloud. The "force" will drive a 2D hand skeleton to "fold" into the 3D joint coordinates representing the hand pose.}
\label{fig:concept}
\end{figure}

With the advancement of deep neural networks (DNNs), various DNN-based hand pose estimation techniques achieved powerful performances. In most of these techniques, 2D convolution neural networks (CNNs) have been adopted to perform direct hand depth image processing \cite{tompson2014real, ge2016robust,guo2017region, ren2019srn, chen2020pose}.  However, 2D CNNs cannot fully take advantage of 3D spatial information of the depth image, which is essential for achieving high accuracy. An intuitive solution is to discretize hand depth images into a 3D voxelized representation and perform 3D-to-3D inference using a 3D CNN \cite{ge20173d, moon2018v2v}. However, its critical limitation is the cubic growth of memory consumption with an increase in the image resolution \cite{riegler2017octnet}. Thus, application of 3D CNNs has been limited to low-resolution images, which may lead to lose of critical details for estimation.

In contrast, the point cloud is being regarded as an efficient and precise representation for 3D hand pose estimation, as it models hand depth images into the continuous 3D coordinates without discretization. However, the point cloud could not be directly processed by conventional DNNs due to the irregular order of points, until the emergence of PointNet \cite{qi2017pointnet}. With a concise symmetric architecture composed of a point-wise shared-weights multi-layer perceptron (MLP) and a max-pooling layer, PointNet is invariant with the order of the input points.

\begin{figure*}
\centering
\includegraphics[width=17cm]{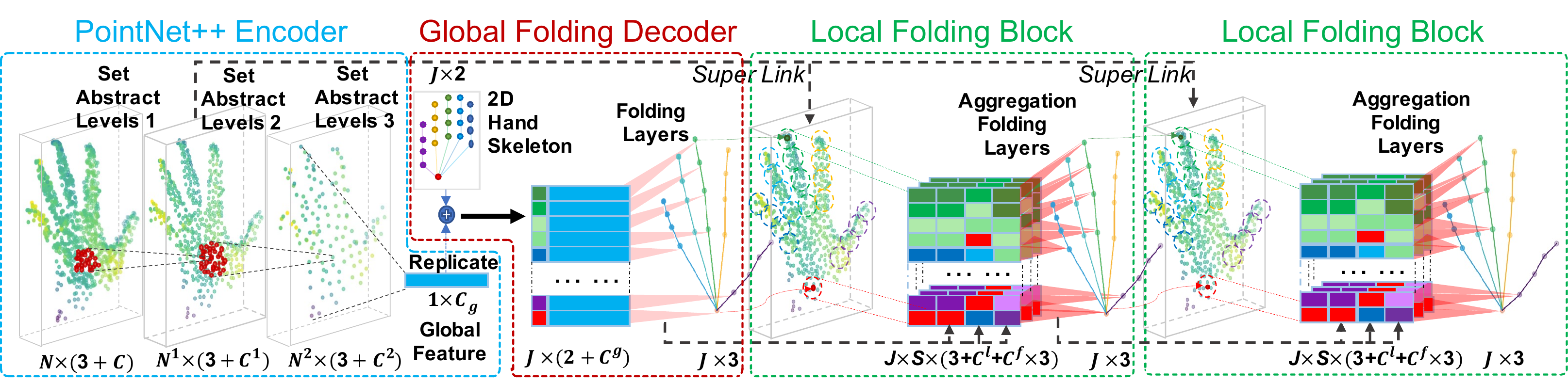}
\caption{The HandFoldingNet architecture. It takes the preprocessed normalized point cloud with surface normal vectors from a 2D depth image as an input. The hierarchical PointNet encoder is then exploited to extract features of various levels to summarize a global feature from the input point cloud. The global folding decoder receives the global feature to guide the folding of a pre-defined 2D hand skeleton into the initial joint coordinates. In the end, the local features near the initial joint coordinates are grouped and fed into the local folding blocks to estimate the accurate joint coordinates.}
\label{fig:architecture}
\end{figure*}

Based on this architecture, a series of PointNet-based hand pose estimation models \cite{ge2018hand, ge2018point, chen2018shpr, li2019point} have been proposed. They can be summarized into two categories: 1) regression-based methods and 2) detection-based methods. Regression-based methods \cite{ge2018hand, chen2018shpr} encode the hand shape into a single global feature through a PointNet-based feature extractor. The global feature representing the hand pose in the high dimensional latent space is fed into a non-linear regression network that performs inference of the joint coordinates. On the other hand, detection-based methods \cite{ge2018point, li2019point} adopt hierarchical features to compute heat-map features for each point. The point-wise features represent the possibility distribution of each joint. However, the existing regression-based and detection-based strategies have limitations. The regression-based methods process only a single global feature, which is not sufficient for highly complex mapping into 3D hand poses. On the other hand, the detection-based methods propagate hierarchical features to each point including the points that contribute little to the specific joint estimation. Therefore, this redundant feature propagation significantly increases the computational cost and slows down the estimation.

To tackle these limitations, we propose HandFoldingNet, an accurate and efficient 3D hand pose estimation network. The key idea of HandFoldingNet is to fold a 2D hand skeleton into the 3D pose, guided by multi-scale features extracted from both global and local information.
The motivation of adopting the folding-based design in FoldingNet \cite{yang2018foldingnet} is that it is suitable for a 3D hand pose estimation task. Essentially, a specific hand pose is a result of applying a force on the human hand skeleton. The folding operation can be interpreted as emulating the "force" applied to the fixed 2D hand skeleton, as shown in Figure \ref{fig:concept}. 
In order to guide folding, HandFoldingNet introduces two novel modules that handle different scales of features: 1) a global-feature guided folding (\textbf{global folding}) decoder and 2) a joint-wise local-feature guided folding (\textbf{local folding}) block. Inspired by FoldingNet, a global folding decoder folds a 2D hand skeleton into the 3D hand joint coordinates. The global feature that guides folding is extracted from the input hand point cloud by a PointNet-based encoder \cite{qi2017pointnet++, ge2018hand, ge2018point}. The local folding block utilizes local features as well as spatial dependencies between the joints, in order to augment joint-wise features and correct the coordinate estimation. Utilization of local features is supposed to compensate for the weakness of conventional regression-based methods. Additionally, unlike the detection-based methods that propagate local features to all the points, we only extract a small region of local features near each joint, in order to avoid massive computations. 

We evaluate our network on ICVL \cite{tang2014latent}, MSRA \cite{sun2015cascaded} and NYU \cite{tompson2014real} datasets, which are challenging benchmarks commonly used for evaluation of a 3D hand pose estimation task. The results show that our network generally outperforms the previous state-of-the-art methods in terms of both accuracy and efficiency. The proposed network achieves the mean distance errors of 5.95mm, 7.34mm and 8.58mm on the ICVL, MSRA and NYU datasets, respectively.
Meanwhile, it contains only 1.28M parameters and runs in real-time with 84 frames per second on a single GPU. 


The key contributions of this paper are as follows:
\begin{itemize}
\setlength{\itemsep}{0pt}
\setlength{\parsep}{0pt}
\setlength{\parskip}{0pt}

\item We propose a novel neural network, HandFoldingNet, which takes the hand point cloud as input and estimates the 3D hand joint coordinates based on the multiscale-feature guided folding. 

\item  We propose a global-feature guided folding decoder that infers joint-wise features and coordinates. The joint-wise features help the model exploit natural spatial dependencies between the joints for better estimation performance.

\item  We propose joint-wise local-feature guided folding to capture local features and spatial dependencies that augments joint-wise features for higher accuracy.

\item  We conduct extensive experiments to analyse the efficiency and accuracy of our proposed network and its key components.

\end{itemize}

\section{Related Work}
\subsection{Depth-based 3D Hand Pose Estimation}


Traditional 3D hand pose estimation approaches based on depth images are mainly implemented in three categories: generative methods \cite{khamis2015learning, tzionas2016capturing, tkach2017online, romero2017embodied}, discriminative methods \cite{keskin2012hand, liang2014parsing}, and hybrid methods \cite{taylor2016efficient, sharp2015accurate, tang2015opening}. In recent years, DNN-based models showed superior performance on 3D hand pose estimation tasks. Representative 2D CNNs are commonly adopted to pose estimation in various implementations. A series of studies \cite{tompson2014real, ge2016robust} exploited 2D CNNs in order to extract a 2D heat-map that represents the possibility distribution of hand joints from a depth image. 

Another line of work proposed regression-based methods based on 2D CNNs \cite{guo2017region, ren2019srn, chen2020pose}, which act as feature extractors that provide efficient features for joint coordinates regression. Instead of processing in the 2D space, several approaches \cite{ge20173d, moon2018v2v} encoded 2D depth images into 3D voxels and adopted 3D CNNs to estimate the 3D hand pose. As depth images can be easily transformed into the point cloud by multiplying the camera intrinsic matrix, several point cloud based models \cite{ge2018hand, ge2018point, chen2018shpr, li2019point} have been proposed. They showed acceptable efficiency and performance by directly processing the input coordinates to estimate the joint coordinates in the identical 3D space.

HandFoldingNet is inspired by these point cloud based methods, but it differs from them in the following aspects. The proposed network does not directly regress the hand joint coordinates nor estimate the point-wise probability distribution. Instead, it first regresses the initial joint coordinates for grouping local features. Meanwhile, it also provides joint-wise features for modeling spatial dependencies. In the end, the network aggregates these local features and spatial dependencies to estimate the accurate joint coordinates. 

\subsection{Deep Point Cloud Reconstruction}

Deep point cloud reconstruction aims to reconstruct the point cloud based on the features extracted from images, point clouds, or other types of data. An intuitive way of achieving the point cloud reconstruction is to adopt 3D CNNs, as in \cite{wu20153d, brock2016generative, girdhar2016learning, sharma2016vconv}. However, these approaches reconstruct the voxelized representation of the point cloud. Instead of CNN-based methods, 
other approaches \cite{achlioptas2018learning, yang2018foldingnet, wang20193dn, cheng2019point} proposed direct reconstruction of the point cloud. 

Theoretically, our main task, estimating hand joint coordinates for a given hand point cloud, can be transformed into the point cloud reconstruction task, because the estimated joint coordinates can be treated as a small set of points that need to be reconstructed. Therefore, we inherit the idea of FoldingNet \cite{yang2018foldingnet} to reconstruct the joint point cloud. FoldingNet proposed a novel folding operation implemented by a sequence of shared-weights MLPs. This folding operation can be intuitively interpreted as learning the "force" to fold a given 2D grid lattice into the target point cloud. There are two critical differences between our network and FoldingNet: 1) we introduce folding of a 2D hand skeleton instead of a regular grid lattice in order to adapt it to the hand pose estimation task, 2) we exploit multi-scale features for higher estimation accuracy, unlike FoldingNet that processes only a single global feature.


\section{HandFoldingNet}

HandFoldingNet aims to perform hand pose estimation using 2D hand joint skeleton folding. The network architecture is shown in Figure \ref{fig:architecture}. It takes an $N \times 6$ matrix $(\mathbf{P}^{nor},\mathbf{F}^{nor})$, which represents a set of normalized points, as an input. Each row of the input matrix is composed of a normalized 3D $xyz$ coordinate $\textbf{p}_i^{nor} \in \textbf{P}^{nor}$ and the corresponding 3D surface normal vector $\textbf{f}_i^{nor} \in \textbf{F}^{nor}$. The output is a $J \times 3$ matrix, representing the 3D coordinates of estimated $J$ joints. The $N$ points are firstly input to the hierarchical PointNet encoder that extracts local features of various levels and a single global feature. Then the global feature is fed into the global-feature guided folding decoder and guides folding of the fixed 2D hand skeleton into the 3D joint coordinates. In order to augment the estimation performance, the output from the global folding decoder and local features near them are processed by joint-wise local-feature based folding blocks.

\subsection{Point Cloud Preprocessing}
First, the 2D depth image is converted into a point cloud by reprojecting the pixels in the 3D space, forming the model input $(\mathbf{P}^{nor},\mathbf{F}^{nor})$.
We follow the point cloud preprocessing method described in HandPointNet \cite{ge2018hand}. The input depth images are first transformed into point cloud representations through camera intrinsic parameters, to adapt to our point cloud based network. Then, in order to deal with various hand orientations, an oriented bounding box (OBB) is created from the 3D point cloud. After that, the point cloud is rotated into the OBB coordinate system, whose axes are aligned with the principle components of the hand points distribution. The oriented points are sub-sampled and normalized into the range of [-0.5, 0.5] to form the final input coordinates $\mathbf{P}^{nor}$. In the end, point-wise surface normal vectors $\mathbf{F}^{nor}$ are calculated from the normalized point cloud. Please refer \cite{ge2018hand} for more details.


\subsection{Hierarchical PointNet Encoder}

\begin{figure*}
\centering
\includegraphics[width=17cm]{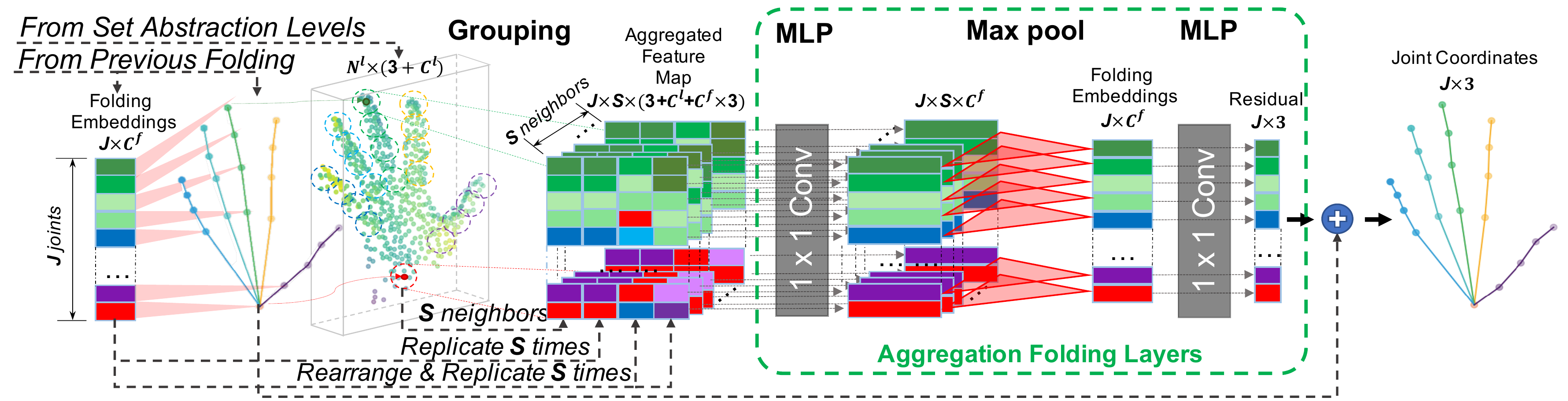}
\caption{Joint-wise local feature guided folding block. The local folding block accepts three inputs, which are the previously estimated joint coordinates, folding embeddings from intermediate layers of the previous folding block, and a local feature map extracted by the previous set abstraction level. The joint coordinates are used as centroids that group local features from the local feature map. Folding embeddings are rearranged to be aligned with the corresponding adjacent joints to collect spatial dependencies. Ultimately, the aggregated feature map composed with grouped local features and rearranged embeddings is fed into a symmetric architecture to compute the residual with respect to the previously-estimated joint locations for more accurate joint estimation.}
\label{fig:lsfaf}
\end{figure*}

We exploit the same hierarchical PointNet encoder as in \cite{ge2018hand, ge2018point} to extract features from the unordered point cloud. 
As shown in Figure \ref{fig:architecture}, the encoder consists of a cascade of $L$ point set abstraction levels. 
The $l$-th level ($l \in \{1, 2,…, L\}$) takes $N^{l-1} \times (3 + C^{l-1})$ matrix from the previous $(l-1)$-th level as an input, of which the $i$-th row is composed of a 3D coordinate $\mathbf{p}_i^{l-1}$ and the corresponding feature $\textbf{f}_i^{l-1}$. 
Then it outputs $N^l \times (3 + C^l)$ matrix, 
which is composed of $N^l$ of sub-sampled centroids $\textbf{p}_i^l$ and their corresponding $C^l$-dim local features $ \mathbf{f}_i^l$.
Specifically, for the first level, the input coordinate is $\textbf{p}_i^{nor}$ and the corresponding feature is a 3D surface normal vector $\textbf{f}_i^{nor}$. 

The $N^l$ centroids are randomly sampled from the input coordinates. Then, $S$ neighbor points with their corresponding features around each centroid $\textbf{p}_i^l$ are gathered as a local region $\{ \textbf{p}_{s,i}^{l-1},\textbf{f}_{s,i}^{l-1} \}_{s = 1}^S$ by using the ball query \cite{qi2017pointnet++} within a specified radius $r$. The coordinates in the local region are then translated to the local frame relative to their centroid: $\textbf{p}_{s,i}^{l-1} -\textbf{p}_i^l$.
For each local region, a symmetric PointNet \cite{qi2017pointnet} with a 3-layer MLP is adopted to generate a $C^l$-dim feature for each point in the region. Subsequently, a max-pooling operation aggregates these point-wise features into a single local feature representing the corresponding centroid. Therefore, the local feature of the $j$-th sub-sampled centroid in the $l$-th level is represented as:
\begin{equation}
\textbf{f}_i^l = \mathop{MAX}\limits_{1 \leq s \leq S} (h([\textbf{p}_{s,i}^{l-1} -\textbf{p}_i^l, \textbf{f}_{s,i}^{l-1}])),
\end{equation}
where $h$ is the MLP, MAX is the channel-wise max-pooling operation, and '[$\cdot$]' is the concatenation operation.

For the last level, it directly adopts the shared-weights MLP and max-pooling operation on the whole input (without sampling) in order to generate the single $C^g$-dim global feature, which is represented as:

\begin{equation}
\textbf{g} = \mathop{MAX}_{1 \leq i \leq N^{L-1}} (h([\textbf{p}_i^{L-1}, \textbf{f}_i^{L-1}])).
\end{equation}

\subsection{Global-Feature Guided Folding Decoder}

\begin{figure}
\centering
\includegraphics[width=4.5cm]{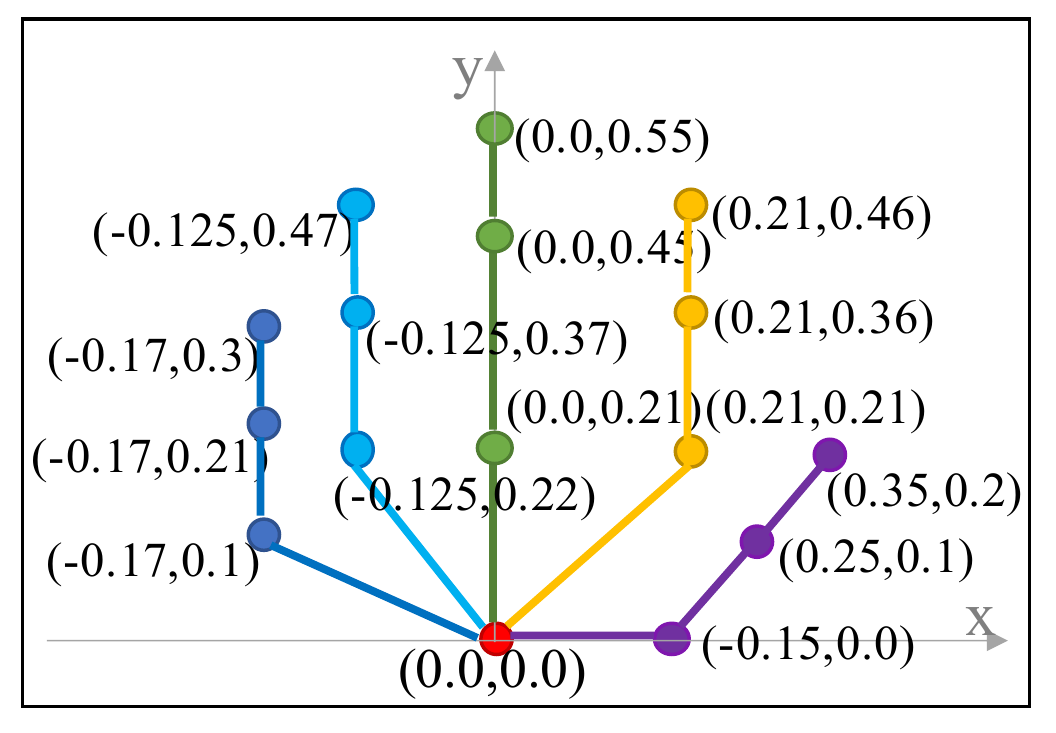}
\caption{An example of a 2D hand skeleton based on the ICVL dataset. The skeleton contains $J=16$ points, each of which is represented as a 2D coordinate.}
\label{fig:grid}
\end{figure}

The proposed decoder folds a fixed 2D hand skeleton into the 3D coordinates of joints, being guided by a global feature. The hand skeleton is a set of hand joint coordinates in a 2D plane and is handcrafted by the following steps: 1) randomly choosing samples from the training set, 2) measuring the average length of links between each pair of adjacent ground truth joints from the samples, 3) unfolding links in a 2D plane, 4) collecting the coordinates of joints across every two connected links. An example of the 2D hand skeleton for the ICVL dataset is shown in Figure \ref{fig:grid}. 

After the hierarchical PointNet encoder extracts the global feature $\textbf{g}$, it is fed into the global folding decoder. Before inserting the global feature $\textbf{g}$, we replicate it $J$ times and concatenate the replicated features with the fixed hand skeleton, whose size is $J \times 2$. The result of the concatenation is supplied to a 2-layer MLP that generates a high-dimensional folding embedding $\mathbf{e_j}$ for each joint. A subsequent 1-layer MLP predicts the initial 3D joint coordinates by processing input embeddings. Hence, the output coordinate $\textbf{j}^0_j$ of the $j$-th joint is represented as:
\begin{equation}
\textbf{j}^0_j = h_p(\textbf{e}_j) \text{  where  } \textbf{e}_j=h_e([\textbf{skel}_j, \textbf{g}]),
\end{equation} 
where $h_p$ and $h_e$ denote the MLPs, $\textbf{e}_j$ denotes the intermediate folding embedding, and $\textbf{skel}_j$ denotes the $j$-th point of 2D coordinate of the fixed skeleton.



\subsection{Joint-Wise Local-Feature Guided Folding Block}
Using only a single global feature (\textit{i}.\textit{e}. global-feature guided folding and other regression-based methods) is not sufficient to accurately estimate the joint coordinates . We believe that the use of additional joint-wise local features encourages the network to correct the joint coordinates.
Therefore, we propose a novel joint-wise local-feature guided folding block for capturing local features and spatial dependencies that help better estimation.

As shown in Figure \ref{fig:lsfaf}, the output coordinates from the $(k-1)$-th folding block are firstly used as centroids for the current $k$-th local folding block. The $J$ centroids group $J$ local regions from the output of the $l$-th set abstraction level within radius $r$. From each region, $S$ neighbors are sampled, each of which is composed of a 3D local coordinate $\textbf{p}_{s,j}^l - \textbf{j}^{k-1}_j$ and a $C^l$-dim corresponding local feature $\textbf{f}_{s,j}^l$, where $1 \leq s \leq S$. 
Therefore, the output size of this grouping is $J \times S \times (3 + C^l)$. 
Note that $l$ is set to 1 as default, while the selection of $l$ will be discussed in Section 4.4.

\begin{figure}
\centering
\includegraphics[width=8cm]{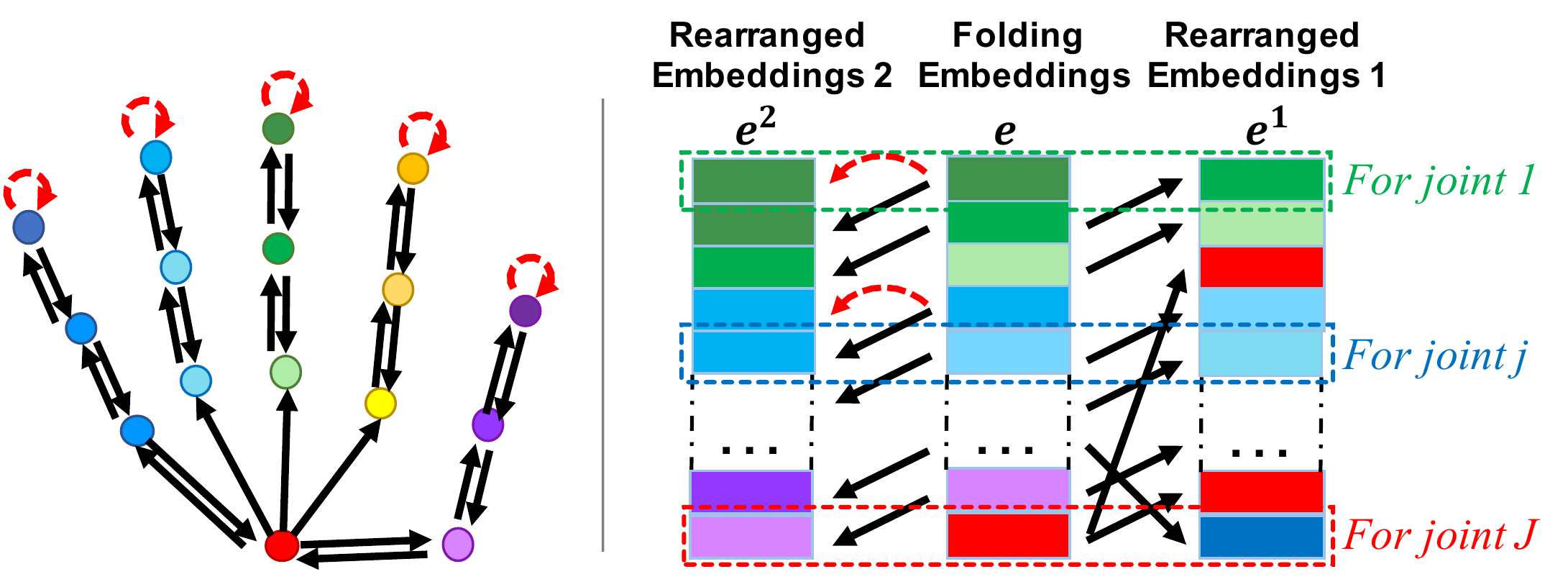}
\caption{The spatial dependency mapping between hand joints of the ICVL dataset (left). Each joint permutes its embedding $\textbf{e}_j$ to map with its two adjacent joints along the mapping direction of the arrows forming two rearranged embeddings $\textbf{e}^1_j$ and $\textbf{e}^2_j$ (right). Exceptionally, fingertips
are forced to map with themselves (red dotted arrows) to keep consistency.}
\label{fig:relation}
\end{figure}

In addition, we introduce a rearrangement process that explicitly models spatial dependencies. It is worth mentioning that, the feature of a specific joint is represented by the corresponding row of the folding embeddings from the global folding decoder. Similarly, the local folding block provides joint-wise folding embeddings as well, enabling the network to stack more local folding blocks for accurate estimation. 
The rearrangement process first permutes the folding embeddings in order to form rearranged embeddings, which match the spatial dependency mapping as shown in Figure \ref{fig:relation}. The $j$-th row of each rearranged embedding is the folding embedding of the adjacent joints of the $j$-th joint. Then, we form the spatial dependency feature map by concatenating rearranged embeddings with the input folding embeddings. 
In the dependency mapping, as shown in Figure \ref{fig:relation}, each joint links with the other two adjacent joints. Therefore, this rearrangement process takes the folding embeddings of size $J \times C^f$ and outputs a spatial dependency map with size $J \times (C^f + C^f + C^f)$. Specifically, since the fingertips only have one adjacent joint, we concatenate them with themselves to keep a uniform shape of the spatial dependency map. As shown in Figure \ref{fig:relation}, there are self-relations for fingertips. 
Moreover, we replicate the spatial dependency feature map $S$ times to align the dimension with the previous grouping output before the following aggregation.

\begin{table}[h!]
\small
\setlength\tabcolsep{5pt}
\begin{center}

\begin{tabular}{c|c|c|c|c|c}
\hline
Block type & r & S & $N^l$ & MLP channels & max\\
\hline
SA ($l$=1) & 0.12 & 64 & 512 & 32, 32, 128 &$\surd$\\
\hline
SA ($l$=2) & 0.2 & 64 & 128 & 64, 64, 256 &$\surd$\\
\hline

SA ($l$=3) & - & 128 & 1 & 128, 128, 512 &$\surd$\\
\hline

global fold ($k$=0) & - & - & J & 256, 256, 3 & $\times$\\
\hline
\multirow{2}{*}{local fold  ($k$=1)} & 0.4 & 64 & J & 256, 256, 256  &$\surd$ \\
\cline{2-6}
  & - & - & J & 256, 256, 3 & $\times$ \\
\hline
\multirow{2}{*}{local fold  ($k$=2)} & 0.4 & 64 & J & 256, 256, 256  &$\surd$ \\
\cline{2-6}
  & - & - & J & 256, 256, 3 & $\times$ \\
\hline

\end{tabular}
\end{center}
\caption{Implementation specifications. Each block contains four types of hyperparameters: search radius (r), the number of grouping neighbors (S), sampling centroids ($N^l$), and the number of output channels of each MLP layer. Max stands for the existence of a max-pooling layer at the end of the block. SA stands for the set abstraction level of PointNet encoder. The local folding blocks are divided into two parts at max-pooling for the clear representation.}
\label{tab:detail}
\end{table}

After local features and the spatial dependency feature map are prepared, we concatenate them together, to form an aggregated feature map. The aggregated feature map is then fed to aggregation folding layers with symmetric structure, as shown in Figure \ref{fig:lsfaf}. In this structure, we introduce a 3-layer MLP and a max-pooling, which aggregate the features into a single folding embedding for each joint. Subsequently, we introduce another 3-layer MLP that maps the high-dimensional embedding into the 3D coordinates. Intuitively, since each joint focuses on its individual local region, only a relative displacement can be effectively computed by this MLP-MAX-MLP structure. Therefore, we inherit the residual block design \cite{he2016deep}. The final joint coordinates are calculated by adding relative displacement outputs with the previously predicted coordinates. Hence, the $j$-th estimated joint of the $k$-th block is represented as:

\begin{equation}
\textbf{j}^k_j = h_r(\mathop{MAX}\limits_{ 1\leq s \leq S}(h_f([\textbf{p}_{s,j}^l -\textbf{j}^{k-1}_j, \textbf{f}_{s,j}^l, \textbf{e}^{k-1}_j]))) + \textbf{j}_j^{k-1},
\end{equation}
where $h_r$ and $h_f$ denote the shared-weights MLPs. $\textbf{j}^{k-1}_j$ indicates the $j$-th output joint coordinate of the previous global folding decoder or local folding block. $\textbf{p}_{s,j}^l$ and $\textbf{f}_{s,j}^l$ are the $s$-th neighbor coordinate and feature of the $j$-th joint where $l$ denotes the $l$-th set abstraction level. 
$\textbf{e}^{k-1}_j$ indicates the concatenation of the $j$-th row of the folding embeddings and its two adjacent joints embeddings from the previous global folding decoder or the local folding block.


\subsection{Loss Function}
As our loss function, we adopt smooth L1 loss, which is less sensitive to outliers than L2 loss. The smooth L1 loss is defined as

\begin{equation}
	L1_{smooth}(\textbf{x}) = \begin{cases}
	0.5|\textbf{x}|, &|\textbf{x}|<0.01\\
	|\textbf{x}|-0.005, &otherwise
		   \end{cases}
    .
\end{equation}

Since the global folding and local folding blocks of our network output their respective estimated coordinates, we supervise all outputs by the following joint loss function:

\begin{equation}
\mathcal L = \sum_{j=1}^J L1_{smooth}(\textbf{j}^0_j - \textbf{j}_j^*) + \sum_{k=1}^K \sum_{j=1}^J L1_{smooth}(\textbf{j}^k_j - \textbf{j}^*),
\end{equation}
where $ \textbf{j}_j^*$ indicates the ground-truth coordinate of the $j$-th joint, and $K$ indicates the quantity of stacked local folding blocks.

\section{Experiments}

\subsection{Experiment Settings}

We conducted experiments on an NVIDIA TITAN RTX GPU with PyTorch. For training, we used the Adam optimizer \cite{kingma2014adam} with beta1 = 0.5, beta2 = 0.999, and learning rate $\alpha$ = 0.001. The number of input points to the network was preprocessed to 1,024 and the batch size was set to 32. The network implementation details are shown in Table \ref{tab:detail}. 
Batch normalization \cite{ioffe2015batch} and the ReLU \cite{nair2010rectified} activation function are adopted in all MLP layers except the layers that output coordinates and residuals. Meanwhile, to avoid overfitting, we adopted online data augmentation with random rotation ([-37.5, 37.5] degrees around z-axis), 3D scaling ([0.9, 1.1]), and 3D translation ([-10, 10]mm). We evaluated the performance of the proposed model using public hand pose datasets, the ICVL \cite{tang2014latent}, MSRA \cite{sun2015cascaded} and NYU \cite{tompson2014real} datasets. We trained the model for 400 epochs on ICVL, 200 epochs on NYU and 80 epochs (with a learning rate decay of 0.1 after 60 epochs) on MSRA.

\subsection{Datasets and Evaluation Metrics}

\begin{table}[h!]
\setlength\tabcolsep{5pt}
\small
\begin{center}

\begin{tabular}{ccccccc}
\hline
\multirow{2}{*}{Methods} & \multicolumn{3}{c}{Mean error (mm)}& \multirow{2}{*}{Input}& \multirow{2}{*}{Type} \\
\cline{2-4}
                        & ICVL          &   MSRA        &   NYU     &   \\
\hline

DeepModel \cite{zhou2016model}       &11.56          & -            & 17.04             &2D   &R  \\
DeepPrior \cite{oberweger2015hands}  &10.4           & -            & 19.73             &2D   &R  \\
Ren-4x6x6 \cite{guo2017region}       &7.63           & -            & 13.39            &2D   &R  \\
Ren-9x6x6 \cite{wan2018dense}        &7.31           & 9.7          & 12.69           &2D   &R  \\
DeepPrior++ \cite{oberweger2017deepprior++}          &8.1           & 9.5           & 12.24           &2D   &R  \\
Pose-Ren \cite{chen2020pose}         &6.79           & 8.65         & 11.81         &2D   &R  \\
DenseReg \cite{wan2018dense}         &7.3            & \textbf{7.2} & 10.2 &2D   &D  \\
CrossInfoNet \cite{du2019crossinfonet}&6.73          & 7.86         & 10.08          &2D   &R  \\
JGR-P2O \cite{fang2020jgr}           &6.02           & 7.55         & \textbf{8.29}          &2D   &D  \\
\hline
3DCNN \cite{ge20173d}                & -             &9.6           & 14.1            &3D   &R  \\
SHPR-Net \cite{chen2018shpr}         &7.22           & 7.76         & 10.78          &3D   &R  \\
HandPointNet \cite{ge2018hand}       &6.94           & 8.5          & 10.54           &3D   &R  \\
Point-to-Point \cite{ge2018point}    &6.3            &7.7           & 9.10            &3D   &D  \\
V2V \cite{moon2018v2v}               &6.28           &7.59          & 8.42           &3D   &D  \\
\hline
Ours                                 &\textbf{5.95}  &\textbf{7.34} & 8.58  &3D   &R\\
\hline
\end{tabular}
\end{center}
\caption{Comparison of the proposed method with previous state-of-the-art methods on the ICVL, MSRA and NYU datasets. Mean error indicates the mean distance error. Input indicates the input representation of 2D (depth image) or 3D (voxel or point cloud). Type D and R indicate the detection-based method and regression-based method, respectively.}
\label{tab:sota}
\end{table}

{\bf MSRA Dataset.} The MSRA dataset \cite{sun2015cascaded} provides more than 76K frames from 9 subjects. Each subject contains 17 hand gestures. The ground truth of each frame contains $J = 21$ joints, including one joint for a wrist and four joints for each finger. Following the most recent work \cite{sun2015cascaded}, we evaluate this dataset with the leave-one-subject-out cross-validation strategy.

{\bf ICVL Dataset.} The ICVL dataset \cite{tang2014latent} is a commonly-used depth stream hand pose dataset that provides 22K and 1.6K depth frames for training and testing, respectively. The ground truth of each frame contains $J = 16$ joints, including one joint for a palm and three joints for each finger. Since the frames also contain the human body area, we firstly crop the hand area from a depth image with the method proposed in \cite{oberweger2017deepprior++}, and take the output joint locations of the global folding decoder to segment the image of the hand area.

{\bf NYU Dataset.} The NYU dataset is captured from three different views. Each view contains 72K training 8K testing depth images captured with the Microsoft Kinect sensor. Following recent works, we only use one view and 14 joints out of total of 36 annotated joints for training and testing.
We also follow the same hand area segmenting process as in the ICVL dataset.

{\bf Evaluation metrics.}
We evaluate the hand pose estimation performance with two commonly-used metrics: the mean distance error and the success rate. The mean distance error measures the average Euclidean distance between the estimated coordinates and ground-truth ones for all the joints over the entire testing set. The success rate is the fraction of the frames whose mean distance error is less than a certain distance threshold.

\begin{figure*}
\centering
\includegraphics[width=17cm]{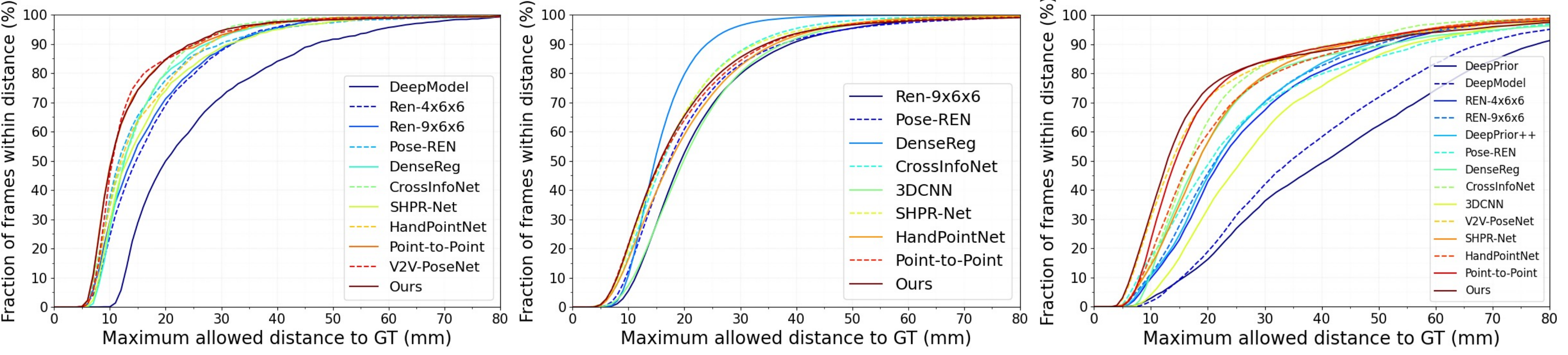}
\caption{Comparison with the state-of-the-art methods using the ICVL (left), MSRA (middle) and NYU (right) dataset. The success rate is shown in this figure.}
\label{fig:threshold}
\end{figure*}

\begin{figure*}
\centering
\includegraphics[width=17.5cm]{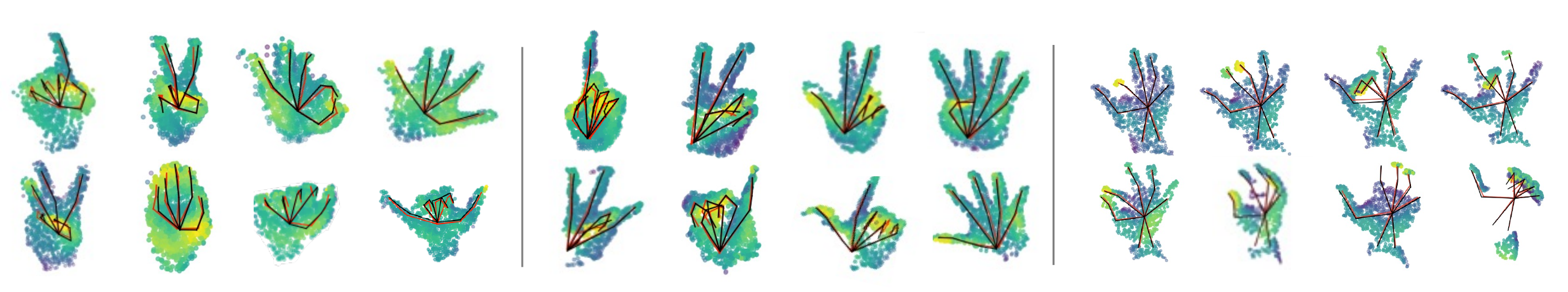}
\caption{Qualitative results of HandFoldingNet on the ICVL (left), MSRA (middle) and NYU (right) dataset. Hand depth images are transformed into 3D points as shown in the figure. Ground truth is shown in black, and the estimated joint coordinates are shown in red.}
\label{fig:qualitative}
\end{figure*}

\subsection{Comparison with State-of-the-arts}

We compare HandFoldingNet with other state-of-the-art methods, including methods with 2D (depth image) input: model-based method (DeepModel) \cite{zhou2016model}, DeepPrior \cite{oberweger2015hands}, improved DeepPrior (DeepPrior++) \cite{oberweger2017deepprior++}, region ensemble network (Ren-4x6x6 \cite{guo2017region}, Ren-9x6x6 \cite{wan2018dense}), Pose-Ren \cite{chen2020pose}, dense regression network (DenseReg) \cite{wan2018dense}, CrossInfoNet \cite{du2019crossinfonet} and JGR-P2O \cite{fang2020jgr}, and methods with 3D (point cloud or voxel) input: 3DCNN \cite{ge20173d}, SHPR-Net \cite{chen2018shpr}, HandPointNet \cite{ge2018hand}, Point-to-Point \cite{ge2018point} and V2V \cite{moon2018v2v}. Figure \ref{fig:threshold} shows the success rate on the ICVL, NYU, and MSRA dataset.
The qualitative results are represented in Figure \ref{fig:qualitative}.

Table \ref{tab:sota} summarizes the performance based on the mean distance error on the three datasets. The results show that our method outperforms the existing methods on the ICVL dataset, achieving the mean distance error of 5.95mm. The proposed model also achieves the second-lowest error on the MSRA dataset and third-lowest error on the NYU dataset. Among methods using the 3D input, our method outperforms other state-of-the-art methods on both ICVL and MSRA datasets. Also, HandFoldingNet shows the state-of-the-art performance among regression-based methods on all three datasets. 
Figure \ref{fig:threshold} represents that our method achieves the highest success rate when the error threshold is lower than 10mm, 13mm and 25mm on the ICVL, MSRA and NYU datasets, respectively.

\subsection{Ablation Study}

We conduct ablation experiments evaluating the performance impact of each component in our model. The following experiments are evaluated based on the ICVL dataset.

\noindent
\textbf{Effectiveness of the local folding block.}
This experiment evaluates the accuracy improvement by attaching the proposed local folding block. To compare with the proposed network having one global folding and two local folding blocks (triple fold), we introduce a shallow network (single fold) that only provides the global folding, a network with only one local folding block (double fold), and a network with three local folding blocks (quadra fold). Table \ref{tab:resfoldblock} shows the performance comparison between the models with different number of local folding.

The result shows that local folding significantly reduces the distance error. This experiment proves that the global folding that only accepts a single global feature for estimation is relatively weak, and the local features contributes the correction of the final joint coordinates. 
Although attaching more local folding blocks increases the inference overhead, the number of parameters and operations of the proposed model (triple fold) are not significant compared to the existing models, as analyzed in Section 4.5.
However, the result also shows that the model performance is saturated at triple fold. 
The reason is that the additional gradients from the third local fold corrupt the back propagation and make the training harder.
Note that double fold still outperforms several point cloud based networks with smaller parameter size and operation count. 




\begin{table}[h!]
\small
\begin{center}

\begin{tabular}{cc|ccc}
\hline

Global & \# Local & Mean  & \multirow{2}{*}{\#Params} &  \multirow{2}{*}{FLOPs} \\
fold & fold &  error (mm)&   &  \\
\hline

$\surd$ & $\times$ &  8.13 & 0.38M & 0.46G\\
$\surd$ & 1        &  6.34 & 0.78M & 0.78G\\
$\surd$ & 2        &  \textbf{5.95} & 1.28M & 1.10G\\
$\surd$ & 3        &  6.08 & 1.78M & 1.48G\\
\hline

\end{tabular}
\end{center}
\caption{Comparison of different numbers of local folding blocks used in the model. \# Local fold indicates the number of local folding blocks attached after the global folding decoder. \# Params indicates the total number of parameters of the network. FLOPs indicates the total number of floating-point operations required for the network inference.}
\label{tab:resfoldblock}
\end{table}

\begin{table}[h!]
\small
\begin{center}

\begin{tabular}{cc|ccc}
\hline

Local & Spatial & Mean & \multirow{2}{*}{\# Params} & \multirow{2}{*}{FLOPs}\\
feature & dependency &  error (mm) &   &  \\
\hline

$\times$ & $\surd$ &  7.90 & 1.21M & 1.04G \\
$\surd$ & $\times$ &  6.35 & 1.08M & 0.91G \\
$\surd$ &  $\surd$ &  \textbf{5.95} & 1.28M & 1.10G \\
\hline

\end{tabular}
\end{center}
\caption{Comparison of different settings between the local feature and spatial dependency. 
}
\label{tab:localspatial}
\end{table}

\begin{table}[h!]
\small
\begin{center}

\begin{tabular}{c|ccc}
\hline

Sampling level &  Mean error (mm) & \#Params & FLOPs \\ 
\hline

input &  6.58 & 1.21M & 1.04G\\
first ($l$=1) &  \textbf{5.95} & 1.28M & 1.10G\\
second ($l$=2) &  6.48 & 1.34M & 1.17G\\
\hline

\end{tabular}
\end{center}
\caption{Comparison of different set abstraction levels for local features. 
}
\label{tab:level}
\end{table}

\begin{table}[h!]
\small
\setlength\tabcolsep{3pt}
\begin{center}

\begin{tabular}{c|cccc}
\hline
Methods         & \# Param & Speed &Time (ms) & GPU Type\\
\hline
V2V-PoseNet \cite{moon2018v2v}     & 457.5M    & 3.5 & 23 + 5.5 & TITAN X\\
HandPointNet \cite{ge2018hand}    & 2.58M     & 48 & 8.2 + 11.3 & GTX1080\\
Point-to-Point \cite{ge2018point}  & 4.3M      & 41.8 & 8.2 + 15.7 & TITAN XP\\
Ours            & \textbf{1.28M}     & \textbf{84}   & \textbf{8.2} + \textbf{3.7} & TITAN RTX\\ 
\hline

\end{tabular}
\end{center}
\caption{Comparison of the model size and inference time for the methods using the 3D input. Speed stands for the frame rate (fps) on a single GPU. Time stands for the total computation time including preprocessing time and model inference time.}
\label{tab:speed}
\end{table}

\noindent
\textbf{Effectiveness of local features and spatial dependencies}.
We evaluate the contribution of the critical feature components of the aggregated feature map, which are the local feature and spatial dependency feature. We conduct two independent experiments: 1) \textit{without local feature} and 2) \textit{without spatial dependency}. For \textit{without local feature}, we remove the grouped local feature component of the aggregated map and maintain the spatial dependency component. For \textit{Without spatial dependency}, we remove rearranged folding embeddings and maintain the local feature. Table \ref{tab:localspatial} shows that the mean distance error increases by 1.55mm without the local features. Similarly, without the spatial dependency, the mean distance error increases by 0.40mm. These experiments show that the both features are critical for improving estimation accuracy. Meanwhile, the local feature contributes to the performance more efficiently, as it requires smaller parameters and FLOPs while achieving better performance than using the spatial dependency.

\noindent
\textbf{Sampling level of local features.}
HandFoldingNet is composed of three set abstraction levels in the PointNet encoder, where each level has different input points density and feature complexity. Therefore, we should carefully determine the abstraction level so that the local folding blocks can effectively collect extra local features.
To analyze the performance impact of the abstraction level, we experiment with the input, first, and second set abstraction levels as the input to the local folding blocks. Table \ref{tab:level} indicates that adopting the output point cloud from the first set abstraction level achieves the highest performance because the neighbor points around the joints are adequate (input points are dense) and the features they provide are effectively informed (input features are complex). On the other hand, the input point cloud is not complex enough as it only includes 3D surface normal vectors.
Consequently, directly using the input point cloud for local folding is not effective in capturing necessary features that can improve the performance. 
Conversely, using higher abstraction level (sampling level 2) degrades the performance. Although the second level features are sufficiently complex, the points are actually sparse in the 3D space. Therefore, the local folding can not group enough points.

\subsection{Runtime and Model Size}

The runtime of HandFoldingNet measured on an NVIDIA TITAN RTX GPU is 11.9ms per point frame in average, including 8.2ms for preprocessing and 3.7ms for network inference. Thus, it can run in real-time at about 84.0fps.
Table \ref{tab:speed} shows our method has the lowest total latency among the 3D-input based methods. Our method also achieves the fastest inference within the point cloud based methods that require 8.2ms of preprocessing time. 
Moreover, the number of parameters of our proposed network is sufficiently small, which is only 1.28M. Compared with previous state-of-the-art models, our model requires the least parameters. 

\section{Conclusion}
In this paper, we proposed HandFoldingNet, a novel and efficient neural network that takes the point cloud as the input and estimates the 3D hand pose. The proposed network achieves the accurate joint coordinates estimation by leveraging the multi-scale features, including the global feature and the joint-wise local feature. Experimental results on three challenging benchmarks showed that our network outperforms previous state-of-the-art methods while requiring the minimal computational resources. Ablation experiments demonstrated the contribution of its key components for better accuracy and efficiency.

\section*{Acknowledgement}
\noindent
This work was partly supported by the Institute of Information and Communication Technology Planning \& Evaluation (IITP) grant on AI Graduate School Program (IITP-2019-0-00421) and ICT Creative Consilience program (IITP-2020-0-00821) funded by the Korea government. Wencan Cheng was supported by the China Scholarship Council (CSC).


\end{document}